\documentclass[letterpaper, 10 pt, conference]{ieeeconf}  % Comment this line out if you need a4paper

\IEEEoverridecommandlockouts                              % This command is only needed if 
\usepackage{graphicx} % for pdf, bitmapped graphics files
\usepackage{amsmath} % assumes amsmath package installed
\usepackage{amssymb}  % assumes amsmath package installed
\usepackage{algorithm2e}
\usepackage{pifont}% http://ctan.org/pkg/pifont
\newcommand{\cmark}{\ding{51}}%
\newcommand{\xmark}{\ding{55}}%
\usepackage{booktabs}
\usepackage{multirow}
\usepackage{cite}
\usepackage{graphicx}
\usepackage{dblfloatfix}
\usepackage{float}
\usepackage{hyperref}
\usepackage[dvipsnames]{xcolor}

\usepackage{xcolor}

\title{\LARGE \bf
Leveraging Self-Supervised Instance Contrastive Learning for Radar Object Detection
}

\author{Colin Decourt$^{1, 2, 3, 4}$, Rufin VanRullen$^{1, 3}$, Didier Salle$^{1, 4}$ and Thomas Oberlin$^{1, 2}$% <-this % stops a space
\thanks{$^{1}$Artificial and Natural Intelligence Toulouse Institute, Université de Toulouse, France
        }%
\thanks{$^{2}$ISAE-SUPAERO, Université de Toulouse, 10 Avenue Edouard Belin, Toulouse 31400, France
       }%
\thanks{$^{3}$CerCO, CNRS UMR5549, Toulouse
       }%
\thanks{$^{4}$ NXP Semiconductors, Toulouse, France}%
}

\begin{document}

\maketitle
\thispagestyle{empty}
\pagestyle{empty}

%%%%%%%%%%%%%%%%%%%%%%%%%%%%%%%%%%%%%%%%%%%%%%%%%%%%%%%%%%%%%%%%%%%%%%%%%%%%%%%%
\begin{abstract}
In recent years, driven by the need for safer and more autonomous transport systems, the automotive industry has shifted toward integrating a growing number of Advanced Driver Assistance Systems (ADAS). Among the array of sensors employed for object recognition tasks, radar sensors have emerged as a formidable contender due to their abilities in adverse weather conditions or low-light scenarios and their robustness in maintaining consistent performance across diverse environments. However, the small size of radar datasets and the complexity of the labelling of those data limit the performance of radar object detectors. Driven by the promising results of self-supervised learning in computer vision, this paper presents RiCL, an instance contrastive learning framework to pre-train radar object detectors. We propose to exploit the detection from the radar and the temporal information to pre-train the radar object detection model in a self-supervised way using contrastive learning. We aim to pre-train an object detector's backbone, head and neck to learn with fewer data. Experiments on the CARRADA and the RADDet datasets show the effectiveness of our approach in learning generic representations of objects in range-Doppler maps. Notably, our pre-training strategy allows us to use only 20\% of the labelled data to reach a similar mAP@0.5 than a supervised approach using the whole training set. 
\end{abstract}

%%%%%%%%%%%%%%%%%%%%%%%%%%%%%%%%%%%%%%%%%%%%%%%%%%%%%%%%%%%%%%%%%%%%%%%%%%%%%%%%
\section{INTRODUCTION}

The automotive industry has recently witnessed a growing interest in integrating radar sensors for Advanced Driver Assistance Systems (ADAS) applications. Radar sensors have garnered attention due to their inherent robustness against adverse weather conditions, cost-effectiveness, and ability to measure the speed and distance of surrounding objects accurately. Despite these advantages, the low resolution of radar sensors and the scarcity of publicly available datasets have somewhat marginalised their use in object recognition tasks. In the last few years, the increasing number of research papers exploring radar as an alternative to traditional camera and LiDAR sensors for such tasks highlights the potential significance of this technology \cite{deepreflecs_ulrich2021,radar_gnn_pcl_fent2023,erasenet,boot_nas_radar,yang2023adcnet,record}.

\begin{figure}[t]
    \centering
    \includegraphics[width=\linewidth]{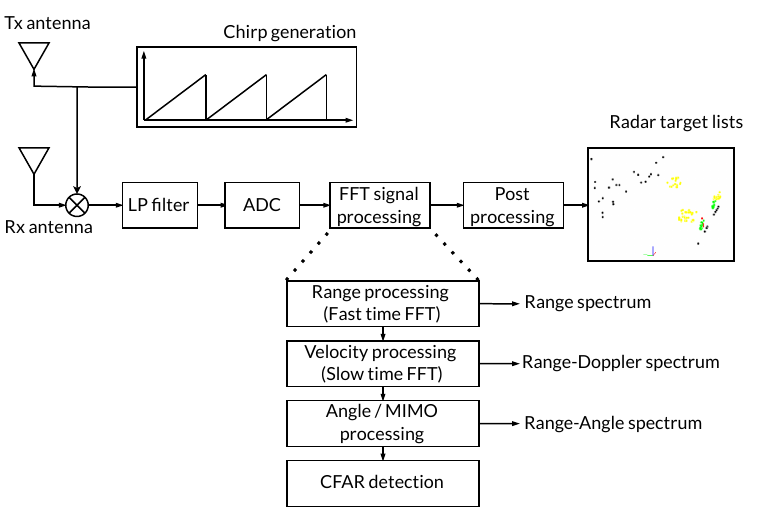}
    \caption{FMCW radar overview}
    \label{fig:radar_dsp}
\end{figure}

Radar data can be represented in two primary forms: point clouds (or target lists) and raw data (or spectra). Target lists provide a condensed representation of detected objects such as the position $(x,y)$, the radial velocity $v_r$, the direction of arrival (DoA) $\theta$ and the radar cross section $\sigma$ of the target. They are obtained after several processing steps as shown in Fig. \ref{fig:radar_dsp}, including signal processing (Fourier transforms, interference mitigation), threshold algorithms (CFAR \cite{blake1988cfar}), and target clustering and tracking (DBSCAN \cite{dbscan} and Kalman filtering \cite{kalman}). Target lists can be used for object recognition tasks like object classification \cite{deepreflecs_ulrich2021,radar_gnn_pcl_fent2023} or segmentation \cite{hybrid_grid_point_detection_ulrich,2d_car_det_pointnets_danzer2019}, but the filtering techniques applied on the received signal might lower the performance of those models. 

On the other hand, raw radar data allows for a more detailed analysis of the received signal, enabling enhanced object discrimination and feature extraction. Several datasets \cite{carrada, raddet, radial,radatron} have enabled research for object recognition tasks on range-Doppler (RD) spectra \cite{darod,object_detection_tracking_rd_yolo_fatseas,radial}, range-azimuth (RA) maps \cite{record,danet_rod2021,major_ra_detection_lstm}, range-azimuth-Doppler (RAD) cubes \cite{mvrss_ouaknine_rad,record,erasenet} and more recently on analog-to-digital-converter (ADC) data \cite{yang2023adcnet,giroux2023_tfftradnet}. This paper focuses on the RD spectrum as it is usually the most efficient representation to use in the radar signal processing chain to detect objects before DoA estimation. 

However, the performance of the approaches mentioned above remains limited by the training datasets' small size and annotation procedures. Frequently, annotation procedures involve a combination of radar, camera, and LiDAR data, introducing complexities and drawbacks such as missing objects. Leveraging the advancements in Self-Supervised Learning (SSL) in computer vision, we aim to address these challenges by proposing a method to reduce reliance on mixed-modality annotations, thereby enhancing the accuracy of radar-based object recognition while reducing the number of labels needed. We propose a novel approach to pre-training object detection models for radar data. The global goal of self-supervised learning is to learn generic representations of the data across tasks (classification, detection, or segmentation). To do so, SSL defines a \textit{pretext task} that allows to learn representations from the dataset without labels \cite{dl_book, ssl_cookbook}. Then, the learned representation can be reused for fine-tuning a model on \textit{downstream tasks} (classification, detection or segmentation). A common pretext task is to encourage two views of the same image to be mapped to similar representations \cite{byol, simclr} (contrastive learning) or to predict missing patches of an image \cite{mae} (image modelling). 
%Self-supervised learning utilises the inherent structure and information within data for training, eliminating the need for externally provided labels.

By using the target lists provided by the radar signal processing chain in Fig. \ref{fig:radar_dsp} and projecting those back on the radar spectra, we can obtain object proposals for free. Object proposals can be used as prior information to learn representations from the dataset without labels, hence in a self-supervised way. Using the proposals from the radar signal processing chain, this paper explores the application of self-supervised learning principles to radar data, specifically focusing on exploiting temporal information and radar detection outputs to pre-train an object detection model for radar. %Our objective is to pre-train object detection models in a self-supervised manner, aiming to reduce the annotation burden associated with radar data while improving the performance of radar object detectors. 
Our contributions are twofold:
\begin{itemize}
    \item First, we propose an instance contrastive learning framework built upon \cite{soco} and \cite{aligndet}, named RiCL (Radar Instance Contrastive Learning), using only range-Doppler maps. The framework is used to pre-train the entire radar object detector, including the backbone, the neck and the head. The model is trained to localise objects and learn object-level representation at multiple ranges. 
    \item Second, we propose a simple and effective method for generating object proposals using the target lists provided by the radar. On a larger scale, this technique will enable easy pre-training using more radar data and reduce the number of annotations required to train radar object detection models.   
\end{itemize}
We conduct experiments and demonstrate significant improvements for the object detection task on RD maps using few labelled data compared to supervised learning. 

Section \ref{sec:ssl_bg} introduces background on self-supervised learning and its limitations for radar. Section \ref{sec:ricl} describes the self-supervised learning framework we propose in this paper. Section \ref{sec:results} presents the results of our experiments on the CARRADA \cite{carrada} and the RADDet \cite{raddet} dataset. Finally, we conclude the paper in Section \ref{sec:conclusion} and present some interesting perspectives.

\section{BACKGROUND ON SELF-SUPERVISED LEARNING} \label{sec:ssl_bg}

\subsection{Self-supervised learning frameworks for images}

There has been a growing interest in SSL for computer vision since 2020 \cite{byol, simclr, mae, convnextv2, deepcluster_caron, dino}, thanks to the availability of large datasets and high-memory GPUs. SSL can be categorised into four families: Deep Metric Learning (DML) \cite{simclr,time_contrastive_learning}, Self-Distillation (SD) \cite{byol,dino,moco,siamsiam}, Canonical Correlation Analysis (CCA) \cite{cca_framework,vicreg,swav}, and Masked Image Modelling (MIM) \cite{mae,convnextv2}. Deep metric learning and self-distillation lie on the concept of contrastive learning \cite{contrastive_loss_idea, contrastive_loss_2}, and aims to encourage the similarity between semantically transformed versions of an input (or \textit{views}). One trains a network to make the embedding of two samples close or far from each other. Generally, because labels are unavailable, different views of the same image are created using image transformations. The CCA family originates from the canonical correlation framework \cite{cca_framework}. The goal of CCA is to infer the relationship between two variables by analysing their cross-covariance matrices, therefore maximising the information content of the embedding. Compared to DML and SD family, CCA does not require large batches or memory bank, momentum encoder or stop gradient operation \cite{vicreg}. Finally masked image modelling consists of randomly masking a high proportion of the input image and learning to reconstruct missing patches. 

\subsection{Pre-training models for object localisation}

In computer vision, SSL frameworks are mostly tuned for image classification and lack good localisation representation. Indeed, the data used for the pre-training contains a single object centred in the image. Zhao \textit{et al.} \cite{dilo} note that due to the transformation applied to the image (random cropping, colour jittering), SSL frameworks are relatively robust to occlusion invariance and tend to learn to use all part of the image to make their predictions. To improve localisation precision of SSL framework, one solution is to rely on carefully chosen unsupervised object priors to learn localised features. Methods propose to modify SSL frameworks with such prior to enhance localisation in their features \cite{soco, densecl, updetr, instance_loc, detreg, videomae, dilo, odin}.

In \cite{densecl, selfpatch}, authors modify the training loss to enforce relationship between extracted features from locations within a single image.  Instead of modifying the loss function, some works explicitly add a prior on object location \cite{soco, instance_loc,aligndet}. This paper adopts a similar strategy by using the detection from the radar as a prior on object location. 

\subsection{Limits of image-based pre-training strategies for radar} \label{subsec:limits_ssl_loc}
Directly applying the methods mentioned above to radar is not possible due to some differences between cameras and radar. First, SSL methods heavily rely on several image transformations (random resizing, cropping, colour jittering) to encourage similarities between two views of the same image. However, almost all of these image transformations cannot be applied to radar due to the characteristics of raw radar data, e.g. complex input, energy loss with range, non-uniform resolution in the angular domain. Second, one key ingredient of SSL is the amount of data. Generally, the more data available to pre-train the model, the more accurate the model on downstream tasks. Most SSL frameworks are pre-trained on large unlabelled image datasets (up to one billion images). However, such large datasets still need to be created for radar and their absence might hamper the benefits of pre-training strategies compared to a fully supervised approach. 

Despite these limitations, recent works explore SSL methods for radar  \cite{ssl_cruw,ssl_radatron}. In \cite{ssl_radatron}, the authors use radar-to-radar and radar-to-camera contrastive loss to learn a general representation from unlabelled RA maps. Nevertheless, this approach relies on camera and on a large dataset compared to the ones used in this paper. In \cite{ssl_cruw}, authors use MIM to pre-train a radar object detector. However, the proposed approach achieves similar performance than a model with a random initialisation, but at lower computational cost. 

In this paper, we show the relevance of pre-training a object detection model (backbone and detection head) in a self-supervised way, using a small dataset. Our work differs from \cite{ssl_cruw} and \cite{ssl_radatron} because we use the detection from the radar to add location prior and we do not rely on any data augmentation techniques.

\section{Radar Instance Contrastive Learning (RiCL)} \label{sec:ricl}

To address the limitations of supervised learning for radar object detection, we propose a novel self-supervised learning framework that pre-trains the entire radar object detection model, including the backbone, neck, and head, without requiring labelled data. We focus on the object detection task, but the representation learned by the model can be applied to other downstream tasks like semantic segmentation or classification. In the context of radar object detection, we leverage temporal information and proposals generated by the radar signal processing chain to provide target data without requiring manual annotations. This approach is particularly valuable for radar data due to the limited availability of labelled data and the challenges associated with manually annotating radar signals. 

Our framework builds upon two existing self-supervised learning approaches for object detection: SoCo \cite{soco} and AlignDet \cite{aligndet}. SoCo focuses on pre-training the backbone network, while AlignDet concentrates on the neck and head of the model. Our approach extends these methods by pre-training the entire radar object detection model, encompassing the backbone, neck, and head. This ensures the model learns a comprehensive representation capable of handling various downstream tasks. Leveraging temporal information and proposals from the radar signal processing chain, our framework effectively learns representations from unlabelled data, overcoming the limitations of supervised learning and expanding the possibilities for radar object detection.

\subsection{Self-supervised learning setup}

\begin{figure*}[t]
    \centering
    \includegraphics[width=0.75\linewidth]{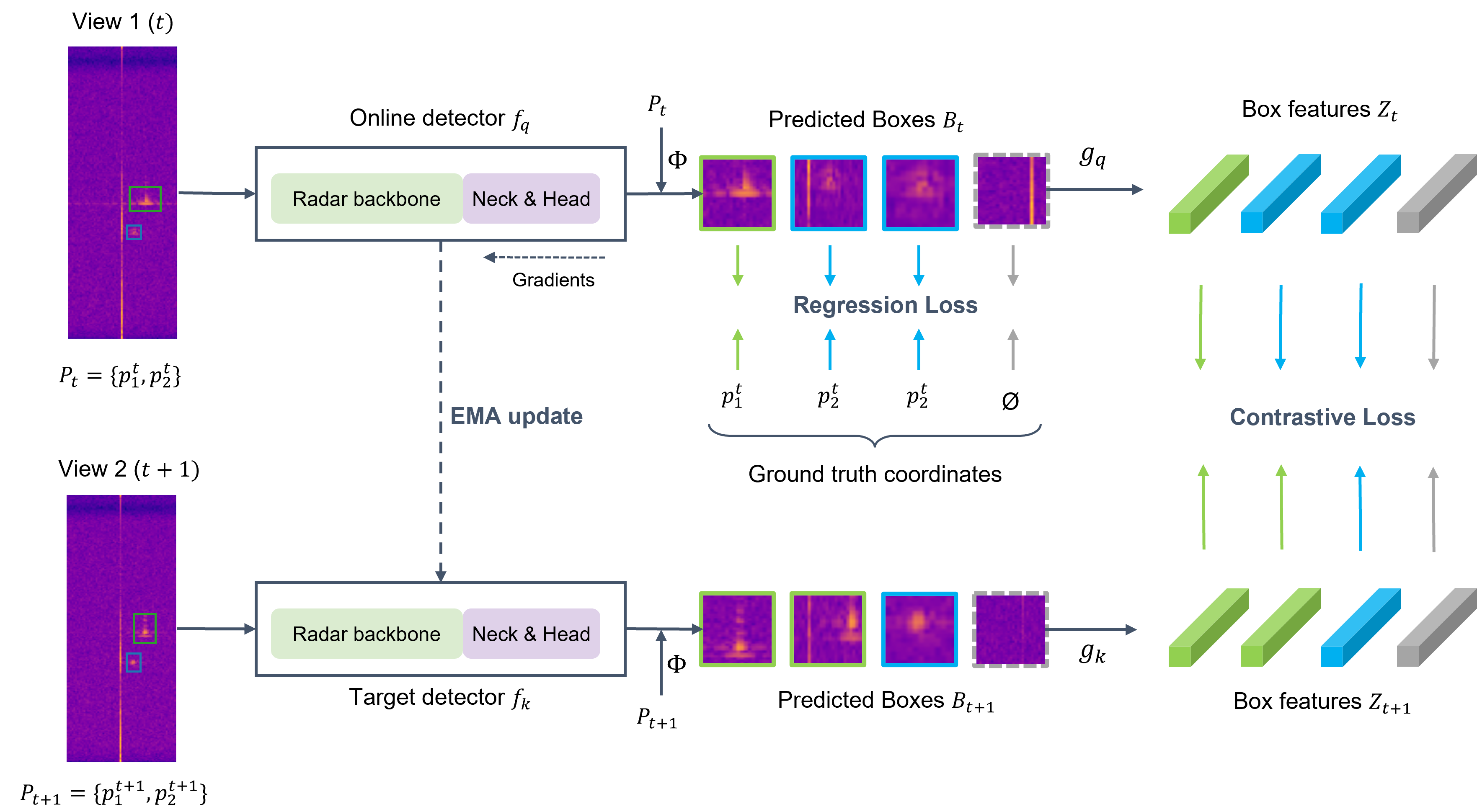}
    \caption{RiCL framework overview. Contrastive learning is performed at the object level, thus maximising the similarity between similar objects at different distance. The regression loss allows the model to learn to localise objects.}
    \label{fig:ricl_overview}
\end{figure*}

Let $r \in \mathbb{R}^{H\times W\times C}$ be an RD spectrum, where $H$ and $W$ are the height and the width of the spectrum and $C$ is the number of channels (\textit{i.e.} antennas) of the radar. Let $P=\{p_1, p_2, \dots, p_n\}$ be a set of proposals obtained from the radar signal processing chain, where each proposal $p_i \in P$ is a bounding box with coordinates $p_i=[x, y, h, w]$. Given a pair of RD maps at two successive time steps $(r_t, r_{t+1})$ and their associated proposals $(P_t, P_{t+1})$, we use two networks (an online network $f_q$ and a target network $f_k$) to predict objects' position. The online and the target networks are two identical CNNs with different weights. Each network receives a different view of the same image, here two successive RD frames, and learns to map one view to another. The online network is updated using gradient descent algorithm, and the target network is updated using an exponential moving average (EMA) of the online network's weights. For each view, we extract object-level features from $f_q$ and $f_k$, project them into a new space using multi-layer perceptron (MLP) and apply a contrastive loss between the features belonging to the same objects in different views. The contrastive loss forces the model's ability to distinguish between different instances of the same object. As proposed in AlignDet, we include a regression loss on the online network to ensure accurate object localisation. Fig. \ref{fig:ricl_overview} summarises the approach.

\subsection{Object proposals generation and matching} \label{subsec:obj_proposal}
We use the radar signal processing chain's output as the basis for contrastive learning. From the radar, we obtain a set $L$ of $N \in \mathbb{N}$ individual reflections (or point clouds) $l_i \in \mathbb{R}^d, i=1, \dots, N$, where $d$ is the number of features. For the frame $r_t$, we refer to the list of reflections as $L_t$. Similarly, we refer to the list of reflections for the frame $r_{t+1}$ as $L_{t+1}$.

For each RD map, if $L_t$ and $L_{t+1}$ are not empty (\textit{i.e.} there are objects in the scene), we cluster the reflections using the DBSCAN \cite{dbscan} algorithm. We obtain a set $C$ of $M \in \mathbb{N}$ clusters such that $\sum_{i=0}^M len(c_i) = N$, where $c_i$ is a set of $m_i \in  \mathbb{N}$ individual reflections. We refer to the set of clusters for the frame $r_t$ as $C_t$. Similarly, we refer to the list of reflections for the frame $r_{t+1}$ as $C_{t+1}$.

For each cluster, we compute statistics $C^{stats}_t, C^{stats}_{t+1}$ like the mean and standard deviation. To create our positive samples for contrastive learning (\textit{i.e.} the same targets at different times), we compute a matrix which contains the pairwise distance between clusters using clusters' statistics:
\begin{equation}
    D_{ij} \mathop{{=}}_{\substack{i \in [1, M_t],\\j \in [1, M_{t+1}]}} dist(C^{stats}_t, C^{stats}_{t+1})
\end{equation}
where $dist$ is the Euclidean distance. Note that other distances can be considered. 

We consider clusters to belong to the same object if the distance between them is lower than a threshold $\epsilon$. Finally, for each valid proposal, we create boxes by projecting the reflections on the range-Doppler map and using minimum and maximum range and Doppler coordinates. Alg. \ref{alg:obj_proposals} summarises the process for each pair. Note that if there is no detection in one of the two pairs, we remove the pair from the dataset used for pre-training the model. 

\RestyleAlgo{ruled}
\begin{algorithm}[t]
\caption{Object proposals generation and matching}\label{alg:obj_proposals}
\SetKwComment{Comment}{/* }{ */}
\KwData{$L_t \in \mathbb{R}^{N_t \times d}$, $L_{t+1} \in \mathbb{R}^{N_{t+1} \times d}$}
\KwResult{Proposals $\{P_t, P_{t+1}\}$}
\eIf{$L_t \neq \emptyset$ and $L_{t+1} \neq \emptyset$}{
    \Comment{If there is a detection}
    $C_t \gets DBSCAN(L_t)$\;
    $C_{t+1} \gets DBSCAN(L_{t+1})$\;
    $(C^{stats}_t, C^{stats}_{t+1}) \gets get\_stats(C_t, C_{t+1})$\;
    $D_{ij} \gets dist(C^{stats}_t, C^{stats}_{t+1})$\;
    $C^{paired} \gets list()$\;
    \Comment{Cluster matching}
    \For{$d_{ij}$ in $D_{ij}$}{
        \If{$d_{ij} < \epsilon$}{
            $C^{paired} \gets append(\{C_t^i, C_{t+1}^j\})$\;
        }
    }
    \Comment{Proposals generation}
    $\{P_t, P_{t+1}\} \gets cluster2boxes(C^{paired})$
}{
    $\{P_t, P_{t+1}\} \gets \{\emptyset, \emptyset\}$\;
}
\end{algorithm}

\subsection{Model pre-training}

We follow AlignDet \cite{aligndet} pre-training strategy, except that we pre-train the entire object detection model and we match objects across successive time steps rather than explicit transformations or augmentations. The following section recalls the AlignDet framework described in \cite{aligndet} that we use to pre-train an object detector.

\paragraph{Self-supervised object detection}
Given an online detector\footnote{Any type of object detector can be used.} $f_q$ and a target detector $f_k$, the online detector and the target detector predict boxes $B_t$ and $B_{t+1}$ from $r_t$ and $r_{t+1}$ such that:
\begin{gather}
    B_t = \phi(h_q^{reg}(f_q^{back}(r_t), P_t) \\
    B_{t+1} = \phi(h_{k}^{reg}(f_k^{back}(r_{t+1}), P_{t+1})
\end{gather}
where $f^{back}$, $h^{reg}$ denote the backbone and the regression head. $\phi$ represents the target assignment operation, which conducts sample matching between $h_q^{reg}(r_t)$ and radar's proposal $P_t$ (\textit{e.g} IoU based assignment). 

In conventional supervised learning, one assigns a category to every matched object in the dataset to compute a classification loss. Here, the index $l \in \{\emptyset, 1, \ldots, n\}$ of each proposal in $P$ is assigned to the paired output of the online detector and target detector as a label. $\emptyset$ stands for the background. 

\paragraph{Box-domain contrastive learning}
We aim to encourage the model to maximise the similarity between two instances of the same object at different time steps. Considering the box representation corresponding to the same object (proposal) should be similar, we employ object-wise contrastive learning under the unsupervised pre-training procedure. 

Considering the predicted boxes $B_t$ and $B_{t+1}$, we extract features of predicted boxes for contrastive learning using an unsupervised classification head $h^{con}$ instead of the classical supervised classification head of the object detector. Then, we project the features using a feature projector $g$ for contrastive learning. For each view, we obtain the following features:
\begin{gather}
    Z_t = g_q(h_q^{con}(r_t, B_t)) \\
    Z_{t+1} = g_{k}(h_k^{con}(r_{t+1}, B_{t+1}))
\end{gather}
where $g$ is a 2-layer MLP head with ReLU (as in AlignDet \cite{aligndet} and MoCoV2 \cite{mocov2}).

\begin{figure*}[ht!]
    \centering
    \includegraphics[width=0.73\linewidth]{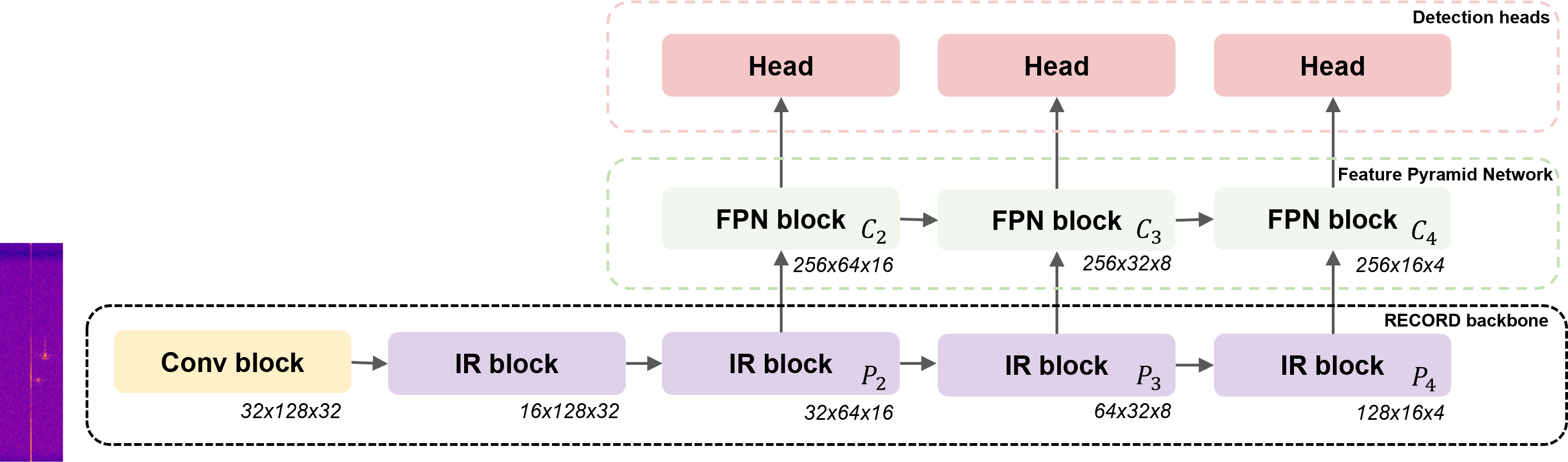}
    \caption{Overview of FCOS model \cite{fcos} using the RECORD backbone \cite{record}, without the LSTMs. IR stands for Inverted Residual bottleneck block as proposed in \cite{mobilenetv2}. Size and number of channels are in $C\times H \times W$ format.}
    \label{fig:fcos_record}
\end{figure*}

Given a query $q\in Q$ where $Q=\{b\in B_t : l \neq \emptyset\}$, assuming its assigned proposal index is $i$ and the feature is $z_q$, we construct sets of positive and negative keys, $Z_+$ and $Z_-$ such that:
\begin{gather}
    Z_+ = \{\ z\in Z_t : l=i \}, \\
    Z_- = \{\ z\in Z_t \cup Z_{t+1} : l\neq i \}
\end{gather}
Then, the box-domain contrastive loss $\mathcal{L}_{con}$ for all query boxes in $Q$ is defined as:
\begin{equation}
    \mathcal{L}_{con} = - \sum_q \sum_{z_+} \log{\frac{\exp{(z_q\cdot z_+/\tau)}}{\exp{(z_q\cdot z_+/\tau)}+\sum_{z_-}\exp{(z_q\cdot z_- / \tau)}}}
\end{equation}

\paragraph{Overall loss}
We optimise the following multi-task loss to pre-train the model:
\begin{equation}
    \mathcal{L} = \lambda_{con}\cdot \mathcal{L}_{con} + \lambda_{reg} \cdot \mathcal{L}_{reg}
\end{equation}
where $\lambda_{con}$ and $\lambda_{reg}$ are the loss hyper-parameters and are chosen with respect to the default setting of the detector, $ \mathcal{L}_{reg}$ is the default regression loss of the chosen object detector like an IoU loss in FCOS \cite{fcos} or an L1-loss in \cite{faster-rcnn,mask-rcnn}. 

\section{EXPERIMENTS} \label{sec:results}

\subsection{Model description}

In this paper, we pre-train the fully convolutional one-stage object detector (FCOS) \cite{fcos}, using the backbone proposed in \cite{record} on which we remove the Bottleneck LSTMs for simplicity and we add a Feature Pyramid Network (FPN)  \cite{fpn} to detect an object at multiple scales. FCOS object detector is a single-stage, anchor-free object detector suitable for automotive applications. As an anchor-free object detector, FCOS does not require tedious hyper-parameters search (\textit{e.g.} anchors, number of proposals), allowing for faster development. This paper focuses only on FCOS object detectors, but RiCL can be applied to other detectors like RetinaNet, Faster R-CNN or YOLO.

Following FPN \cite{fpn}, we detect different sizes of objects on different levels of feature maps. We use three levels of feature maps, defined as $P_2$, $P_3$ and $P_4$. $P_2$, $P_3$ and $P_4$ are produced by backbone CNN's feature maps $C_2$, $C_3$ and $C_4$ followed by a $1\times 1$ convolutional layer with the top-down connections in \cite{fpn}. Fig. \ref{fig:fcos_record} describes the architecture of the backbone and the model. $P_2$, $P_3$ and $P_4$ have strides $4, 8$ and $16$ respectively.

FCOS does not use anchors to detect objects. Instead, for each foreground pixel at location $(x, y)$, FCOS predicts the distances from the location to the four sides of the bounding box. This results in a vector $\mathbf{t^*}=(l^*, t^*, r^*, b^*)$. Each level is responsible for detecting objects of different sizes. In this paper, $P_2$ is responsible of detecting objects in size range $(0, 16)$, $P_3$ in range $(16, 32)$ and $P_4$ in range $(32, \infty)$. 

We use the same parameters as in the FCOS paper \cite{fcos} for the detection head. All the experiments are conducted using the MMDet framework\footnote{https://github.com/open-mmlab/mmdetection}.

\subsection{Pre-training settings}

\paragraph{Dataset and object proposal generation}
We pre-train the model using the range-Doppler maps of the training set of the CARRADA dataset \cite{carrada}. We generate object proposals for each frame using the algorithm presented in Section \ref{subsec:obj_proposal}. For the DBSCAN algorithm, we consider two objects belonging to the same cluster if their Euclidean distance is lower than 2.5, and a cluster is valid if there are more than four reflections. The $\epsilon$ value to match clusters in successive frames is set to $4$. All these values were found empirically and have been shown to provide the best object proposal generation.  

\paragraph{Optimisation settings}
We pre-train the model using the AdamW \cite{AdamW} optimiser, with a learning rate of $1e^{-4}$ and a weight decay of $1e^{-3}$. During pre-training, we use a Cosine Annealing learning rate scheduler \cite{cosinelr} with the default PyTorch hyper-parameters. We pre-train FCOS under the RiCL framework for 12 epochs only and set the batch size to 32. We also use data augmentation, such as random horizontal flipping with a probability of 50\%, to increase the dataset's diversity.

\subsection{Finetuning settings}
\begin{table*}[!t]
\caption{mAP@0.5 on the CARRADA and RADDet test sets for different amounts of labelled data. 100\% stands for training on the entire training set. Pre-training is done on the CARRADA dataset. We use a random initialisation strategy for the models that are not pre-trained. Results are averaged over five different folds. Best results are in bold.}
\label{tab:results}
\centering
\resizebox{\linewidth}{!}{%
\begin{tabular}{cccccccc}
\hline
\multirow{2}{*}{Model} & \multirow{2}{*}{Dataset} & \multirow{2}{*}{Pre-trained} & \multicolumn{5}{c}{Training labelled data}                                             \\ \cline{4-8} 
                          &                          &                              & 100 \%        & 50 \%         & 20\%          & 10 \%         & 5 \%          \\ \hline
FCOS             & \multirow{2}{*}{CARRADA} & \xmark                       & 44.5 $\pm$ 2.5  & 41.7 $\pm$ 1.11 & 38.9 $\pm$ 1.5  & 34.6 $\pm$ 1.5  & 22.9 $\pm$ 9.31 \\
FCOS + RiCL &
   &
  \cmark &
  \textbf{49.2 $\pm$ 0.9 \textcolor{ForestGreen}{(+ 4.6)}} &
  \textbf{49.1 $\pm$ 1.4 \textcolor{ForestGreen}{(+ 7.4)}} &
  \textbf{44.0 $\pm$ 1.7 \textcolor{ForestGreen}{(+ 5.8)}} &
  \textbf{42.9 $\pm$ 1.5 \textcolor{ForestGreen}{(+ 8.3)}} &
  \textbf{40.1 $\pm$ 1.5 \textcolor{ForestGreen}{(+ 17.2)}} \\ \hline
FCOS             & \multirow{2}{*}{RADDet}  & \xmark                       & 39.7 $\pm$ 0.01 & 32.1 $\pm$ 0.02 & 18.8 $\pm$ 0.08 & 15.3 $\pm$ 0.06 & 14.2 $\pm$ 0.01 \\
FCOS + RiCL &
   &
  \cmark &
  \textbf{41.0 $\pm$ 0.01 \textcolor{ForestGreen}{(+ 1.3)}} &
  \textbf{35.1 $\pm$ 0.01 \textcolor{ForestGreen}{(+ 3.0)}} &
  \textbf{25.9 $\pm$ 0.02 \textcolor{ForestGreen}{(+ 7.1)}} &
  \textbf{21.8 $\pm$ 0.02 \textcolor{ForestGreen}{(+6.5)}} &
  \textbf{17.6 $\pm$ 0.005 \textcolor{ForestGreen}{(+ 3.4)}} \\ \hline
\end{tabular}%
}
\end{table*}
\paragraph{Datasets}
We finetune FCOS on the CARRADA \cite{carrada}, and the RADDet \cite{raddet} datasets. When finetuned on the RADDet dataset, we use the weights learned on the CARRADA dataset. 

We use the training, validation and test set provided by the authors of the CARRADA dataset to finetune the model on the CARRADA dataset. For RADDet, we use 90\% of the training set for training and 10\% for validation. The default test set of RADDet is used to assess performance.

To investigate the impact of the number of available labelled data on performance under a finetuning setting, we randomly select 50\%, 20\%, 10\% and 5\% of the training set. We use k-fold cross-validation to provide more reliable results, with $k=5$. 

\paragraph{Optimisation settings}

On the CARRADA dataset, we finetune the model using the SGD optimiser, with a learning rate of $1e^{-2}$ for 20 epochs. We reduce the learning rate by $10$ at epochs 15 and 19. The batch size is set to 32 for all experiments, except when we use 5\% of the training set, where the batch size is set to 4. 

On the RADDet dataset, we finetune the model using the AdamW \cite{AdamW} optimiser, with a learning rate of $1e^{-4}$ and a weight decay of $1e^{-3}$. We use a Cosine Annealing learning rate scheduler \cite{cosinelr} with the default PyTorch hyper-parameters. We use the same batch size as on the CARRADA dataset. 

We use the same parameters for finetuning and training the model using supervised learning. 

\paragraph{Evaluation metric}
We evaluate all the models using the mAP@0.5, a well-known metric for object detection. 

\subsection{Results}

Tab. \ref{tab:results} shows the results of the FCOS object detector, pre-trained or not, on the CARRADA and the RADDet datasets. 
\paragraph{Results using all the labelled data}
When trained with all the labelled data, pre-training the backbone and the detection head using RiCL improves the mAP@0.5 by 4.6 points on the CARRADA dataset and 1.3 points on the RADDet dataset, compared to a random initialisation approach. Surprisingly, despite being different datasets, the representations learned on the CARRADA dataset transfer well on the RADDet dataset. Therefore, the results in Tab. \ref{tab:results} confirm the effectiveness of RiCL in learning the generic representation of objects in an unsupervised way. Using information from successive frames helps the model learn more diverse representations of the same objects (\textit{i.e.} a car, a pedestrian). 

We notice RiCL leads to a better improvement using 100\% of the data on the CARRADA dataset than on the RADDet dataset. Indeed, because we pre-train FCOS on the CARRADA dataset, the model has already seen data from the training dataset. Therefore, the model is in a better initialisation state. However, the validation and the testing set being unseen, the comparison with a random initialisation remains fair. It shows the relevance of a self-supervised learning strategy to pre-train radar object detectors. Moreover, assuming that we want to reduce the annotation work for radar data and generalise RiCL to a large-scale dataset, one might use the same dataset (\textit{i.e.} same radar, similar scenarios) to pre-train and finetune detection model as proposed \cite{ssl_radatron}. 

\paragraph{Impact of the number of labelled data}
The ultimate goal of the proposed approach is to avoid the laborious annotation procedure of radar data; we evaluate the impact of the number of available labelled data on performance. This way, we \textit{simulate} a smaller dataset. We train (starting from random initialisation) and finetune (after pre-training with RiCL) FCOS using 50\%, 20\%, 10\% and 5\% of the training set. Overall, the RiCL pre-training strategy improves the detection performance under a low-data regime, particularly on the CARRADA dataset. 

On the CARRADA dataset, results in Tab. \ref{tab:results} show that using a proper self-supervised pre-training strategy like RICL enables higher mAP@0.5 with half of the labelled data compared to a random initialisation approach. Moreover, RiCL allows the use of only 20\% of the labelled data to reach the same performance level as a fully supervised approach (100\% of labelled data, random initialisation). Finally, we find that the drop in performance as the number of available data decreases is smaller for a model pre-trained with RiCL than for a non-pre-trained model.

On the RADDet dataset, we note that improving pre-training using RiCL is less significant than on the CARRADA dataset. This might be due to the distribution shift between those datasets. In practice, one may use data from the same sensors to pre-train the model, as done in \cite{ssl_radatron}. However, Tab. \ref{tab:results} shows RiCL relevance in boosting detection performance under a low-data regime, particularly when 20\% and 10\% of the labelled data are used. 

\section{CONCLUSIONS AND PERSPECTIVES} \label{sec:conclusion}
This paper proposed a novel approach to pre-train radar object detectors. Though our approach is built upon existing frameworks \cite{soco,aligndet}, we show that combining those and leveraging the outputs of radar sensors allows us to learn generic representations of objects on RD data in a self-supervised way. We propose an object proposal generation method using the target lists from the radar, allowing easy and efficient pre-training of any object recognition tasks using raw radar data. Experiments on different datasets show the relevance of our approach in boosting detection performance.

%This work focused on showing the effectiveness of RiCL in improving the performance of radar object detection models under a low-data regime. Therefore, we did not perform a hyperparameter search to improve the model's performance in this work. % Thomas: supprimer ce paragraphe? Je vois pas trop l'intérêt... Ou alors le tourner plus positif : "in the future we intend to test this method at larger scale, and improve the resutls by using more involved architectures and hyperparameter search"

Although RiCL improves the performance of FCOS significantly, we did not compare it with conventional pre-training strategies used in computer vision like BYOL \cite{byol}, MoCo \cite{moco} or SimCLR \cite{simclr}. Indeed, these approaches rely on numerous data augmentation strategies that can not be applied to raw radar data. Instead, we intend to compare RiCL with the radar pre-training strategy called Radical \cite{ssl_radatron} as soon as the code will be available, or with a multi-frame pre-training strategy such as VideoMoCo \cite{videomoco} or VideoMAE \cite{videomae}. %Thomas: intérêt de la dernière phrase?

Finally, we use the CARRADA dataset to demonstrate the relevance of RiCL because of its simplicity and the availability of radar reflections. However, one key ingredient of self-supervised learning is the amount of data, and the CARRADA dataset remains small. In the future we aim at testing this method at larger scale, and improve the results by using more involved architectures and hyper-parameter search.
% Je supprimerais toute la suite, il vaut mieux éviter l'aut-flagellation :-). Tu peux dire à la place qch comme la phrase suggérée ci-dessus
%For example, authors of \cite{ssl_radatron} use 152k frames to pre-train their model, while we use only 8k frames. Nevertheless, the Radatron \cite{radatron} dataset being too large, we were not able to download it. We consider using the RADIal dataset \cite{radial} (around 30k frames) in future works to validate our approach on a larger dataset and high-resolution data. 

We hope this work advances research in self-supervised learning for radar object recognition. Although we experimented on the object detection tasks, this work benefits other object recognition tasks such as semantic segmentation or classification. Radar provides easy pseudo-labelling, and the size of public radar datasets is growing. Therefore, we believe self-supervised learning with the right pretext tasks and data will help develop radar foundation models, hence improving radar perception systems.

% Globalement je serais pour raccourcir et restructurer la conclusion

\bibliographystyle{IEEEtran.bst}
\bibliography{IEEEfull}

\end{document}